\documentclass[10pt,twocolumn,letterpaper]{article}

\usepackage{cvpr}
\usepackage{times}
\usepackage{epsfig}
\usepackage{graphicx}
\usepackage{amsmath}
\usepackage{amssymb}
\usepackage{booktabs}

\usepackage{algorithm,algpseudocode}

\usepackage[bookmarksnumbered=true]{hyperref} 
\hypersetup{
     colorlinks = true,
     linkcolor = blue,
     anchorcolor = blue,
     citecolor = blue,
     filecolor = blue,
     urlcolor = blue
     }

\cvprfinalcopy % *** Uncomment this line for the final submission

\def\cvprPaperID{****} % *** Enter the CVPR Paper ID here

\usepackage{comment}

\begin{document}

%
%%%%%%%%% TITLE
% \title{Tell Me Where to Spike: Looking Deeper in Spiking Neural Networks}
\title{Visual Explanations from Spiking Neural Networks using Interspike Intervals}

\author{Youngeun Kim\\
Department of  Electrical Engineering\\
Yale University\\
{\tt\small youngeun.kim@yale.edu}
\and
Priyadarshini Panda\\
Department of  Electrical Engineering\\
Yale University\\
{\tt\small priya.panda@yale.edu}
% \\
% \\
% \hspace{-7cm} *UNDER REVIEW. DO NOT DISTRIBUTE.
}
\maketitle
%\thispagestyle{empty}
%%%%%%%%% ABSTRACT
\begin{abstract}
   Spiking Neural Networks (SNNs) 
   compute and communicate with asynchronous binary temporal events that can lead to significant energy savings with neuromorphic hardware. Recent algorithmic efforts on training SNNs have shown competitive performance on a variety of classification tasks. However, a visualization tool for analysing and explaining the internal spike behavior of such temporal deep SNNs has not been explored. In this paper, we propose a new concept of bio-plausible visualization for SNNs, called Spike Activation Map (SAM). The proposed SAM circumvents the non-differentiable characteristic of spiking neurons by eliminating the need for calculating gradients to obtain visual explanations. Instead, SAM calculates a temporal visualization map by forward propagating input spikes over different time-steps. SAM yields an attention map corresponding to each time-step of input data by highlighting neurons with short inter-spike interval activity. 
    Interestingly, without both the backpropagation process and the class label, SAM highlights the discriminative region of the image while capturing fine-grained details. With SAM, for the first time, we provide a comprehensive analysis on how internal spikes work in various SNN training configurations depending on optimization types, leak behavior, as well as when faced with adversarial examples.
    % we compare the two representative SNN optimization methods: conversion and surrogate gradient training.
    % Moreover, we demonstrate that the proposed SAM is robust on the adversarial attack.
    % This new visualization tool may be helpful to the future SNNs development as well as theoretical neuroscience studies.
\end{abstract}

%%%%%%%%% BODY TEXT

\vspace{-2.5mm}
\section{Introduction}
\vspace{-1mm}
% -- Before edit (P1: ANN - explainability)
% Artificial Neural Networks (ANNs) have been achieved human-level performances in various computer vision fields. 
% Despite the performance is increasing, explainability of ANNs are decreased as the number of parameters in networks are skyrocketed.
% To address the problem, recent studies have been proposed a visualization tool such as CAM and Grad-CAM , which represents the most discriminative part of the image.
% Through this work, the networks can enhance the its ability on localization, segmentation, and visual question answering.

% P1: introduce SNN
% For gradcam \cite{selvaraju2017grad,lee2020spike,calabrese2019dhp19,amir2017low, zhou2016learning,deng2009imagenet,wu2018group,reich2000interspike,goodfellow2014explaining,richman2000physiological}

Artificial Neural Networks (ANNs)  \cite{he2016deep,szegedy2015going,iandola2016squeezenet} have shown human-level performance on a wide variety of tasks but incur huge computational cost.
%In the past decade, ANNs have become increasingly complex with more number of parameters and multiply-and-accumulate (MAC) operations to achieve human-level performance.
For instance, while ResNet-50 \cite{he2016deep} reduces the top-5 error by 11.1\% on ImageNet dataset compared to AlexNet, it requires about 5$\times$ more energy for classifying one image \cite{sze2017efficient}.
However, in many real-world applications, neural networks are  required to be implemented on resource-constrained platforms.
Spiking Neural Networks (SNNs) \cite{roy2019towards,panda2020toward,cao2015spiking,diehl2015unsupervised,comsa2020temporal} offer an alternative way for enabling low-power artificial intelligence.
SNNs emulate biological neuronal functionality by processing visual information with binary events (\ie, spikes) over multiple time-steps.
This discrete spiking behavior of SNNs have been shown to yield high energy-efficiency on emerging neuromorphic hardware \cite{furber2014spinnaker,akopyan2015truenorth,davies2018loihi}.

% \begin{figure}
%     \centering
%     \includegraphics[width=0.45\textwidth]{figures/ann_vs_snn.png}
%     \vspace{-2mm}
%     \caption{ 
%     The illustration of difference between Artificial Neural Networks (ANNs) and Spiking Neural Networks (SNNs).
%     Different from the ReLU activation in ANNs that can compute the exact backward gradient,
%     Leaky-Integrate-and-Fire (LIF) neurons in SNNs are non-diffentiable.
%     Therefore, ANNs can calculate the accurate contribution of neuron in previous layer (green) with respect to the specific neuron (red).
%     On the other hand, SNNs are hardly obtain the exact backward gradient. 
%     % The discrepancy between original gradient and surrogate gradient functions induces inaccurate gradient value. In contrast, proposed Spike Activation Map (SAM) is based on bio-plausible Inter-Spike-Interval (ISI), thereby obtains meaningful information. Here, spikes focus on the edge of the object.
%     }
%     \label{fig:motivation}
%     \vspace{-3mm}
% \end{figure}

% P2: SNN optimization
% Due to the growing needs for low-power neuromorphic computing, how to train deep SNNs remains a challenging issue.
% This challenge came from the non-differentiable functionality of an IF (or LIF) neuron (see Section X).

Optimization methods for SNNs have made great strides on image classification tasks over the recent past. Conversion methods \cite{sengupta2019going,han2020rmp,diehl2015fast,rueckauer2017conversion} convert a pre-trained ANN to an SNN by normalizing firing thresholds or weights to transfer ReLU activation to Integrate-and-Fire (IF) spiking activity. So far, conversion techniques have been able to achieve competitive accuracy with ANN counterparts on large-scale architectures and datasets but incur large latency or time-steps for processing.
On the other hand, surrogate gradient descent methods
\cite{lee2016training, han2020rmp,kim2020revisiting} train SNNs using an approximated gradient function to overcome the non-differentiability of the Leaky-Integrate-and-Fire (LIF) spiking neuron \cite{izhikevich2003simple}.
Such methods enable SNNs to be trained from scratch with lower latency on conventional deep learning frameworks (\eg, TensorFlow \cite{abadi2016tensorflow}) with reasonable classification accuracy.

{
Despite improvement in optimization techniques, there is a lack of understanding pertaining to internal spike behavior of SNNs compared to conventional ANN.
% \textit{What is the role of each layer for propagating temporal spikes in SNNs?} 
%\textit{What is the characteristic of visual explanations in SNNs across different time-steps for a given prediction?}. 
Neural networks have been conceived to be ``black-boxes". However, with ubiquitous usage of neural networks, there is a need to understand what happens when a network predicts or makes a decision.  %underlying mechanism for why the networks are making such decision becomes important with increasing the internal complexity of networks. 
On the ANN front, several interpretation tools have been proposed \cite{vondrick2013hoggles,dosovitskiy2016inverting,zintgraf2017visualizing,selvaraju2017grad}  and have found practical usage for obtaining visual explanations and understanding the network prediction.
On similar lines, an SNN interpretation tool is also highly crucial because low-power SNNs are increasingly becoming viable candidates for deployment in real-world applications such as medical robots \cite{bing2018survey}, self-driving cars \cite{hwu2017self}, and drones \cite{salt2019parameter}, where explainability in addition to performance is critical.
In this work, we aim to shed light on the explainability of SNNs.
}

\begin{figure}
    \centering
    \includegraphics[width=0.48\textwidth]{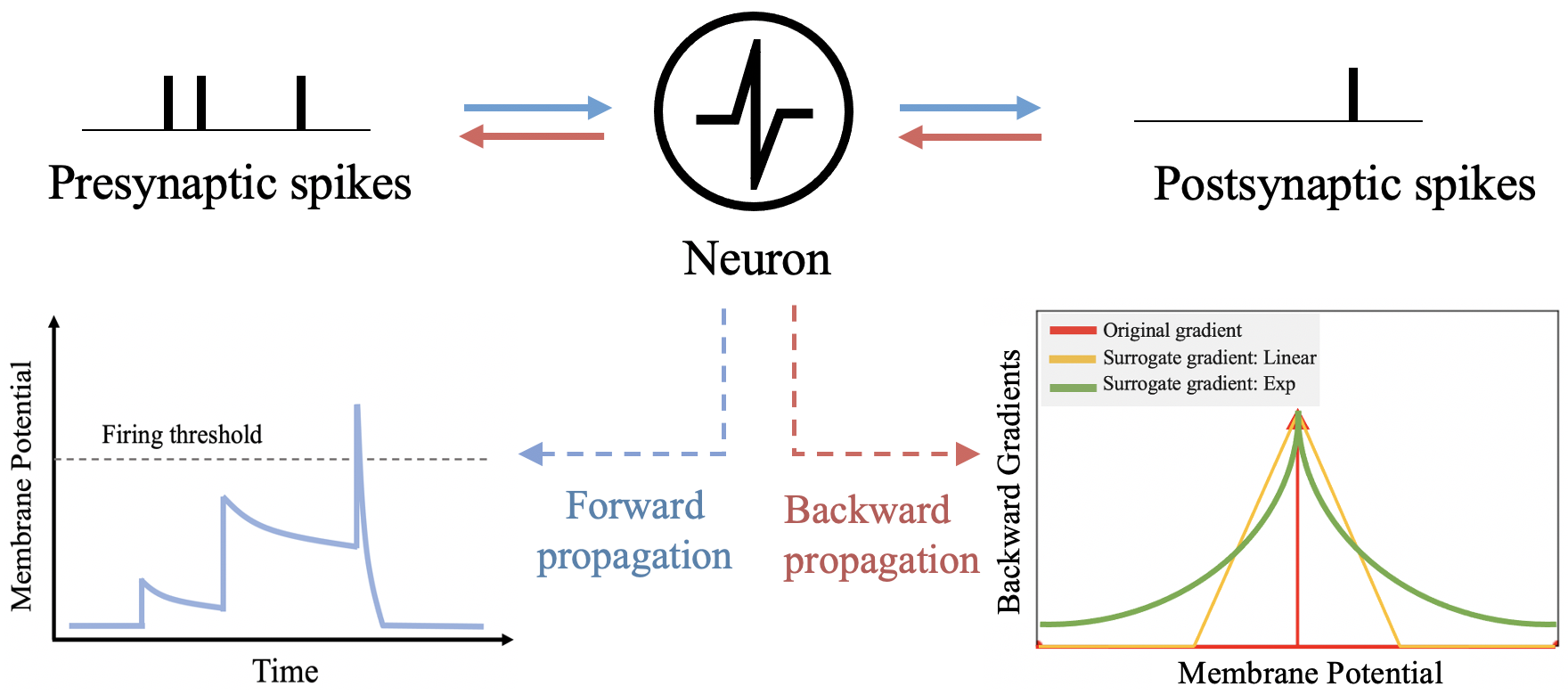}
    % \vspace{-3mm}
    \caption{ 
    The illustration of forward propagation (blue arrow) and backward propagation (red arrow) of an  LIF neuron. 
    During forward propagation, the membrane potential increases according to the pre-synaptic spike input. If the membrane potential exceeds the firing threshold, the LIF neuron generates the post-synaptic spike with the membrane potential reset. This integrate-and-fire behavior induces the non-differentiability of the membrane potential. Therefore, surrogate gradient functions are used to implement the backward gradient.   
    }
    \label{fig:neuron_dynamic}
    \vspace{-6mm}
\end{figure}

The na\"ive approach for explainability is to exploit widely used visualization tools from ANN domain.
Among them, Grad-CAM \cite{selvaraju2017grad} has a huge flexibility in terms of application, and is also used by state-of-the-art interpretation algorithms \cite{hohman2018visual}. 
The authors of Grad-CAM show that the contribution of a neuron from shallow layers to deep layers towards any target class can be quantified by calculating the gradient with backpropagation.
But, SNNs cannot compute exact gradient (\ie, contribution) because of the non-differentiable integrate and firing behavior of an LIF neuron (see Section \ref{sec: background}) as shown in Fig. \ref{fig:neuron_dynamic}.
Therefore, a new concept of visualization for SNNs is required.

In this study, we propose a novel visualization tool for SNN, called Spike Activation Map (SAM), which does not require any backpropagation. Instead, we calculate an attention map by monitoring neurons that carry more information (\ie, spikes) over different time-steps during forward propagation. We exploit the biological observation that short Inter-Spike-Interval (ISI) spikes have more information in a neurological system
\cite{reich2000interspike,snider1998burst,shih2011improved} because these spikes are more likely to induce post-synaptic spikes by increasing the membrane potential of the neuron.
Specifically, for each neuron, we compute a \textit{neuronal contribution score} (NCS) for the prediction.
The NCS score is defined as the sum of \textit{temporal spike contribution score} (TSCS) with an exponential kernel.
The TSCS score assigns high value for spikes firing within a short time window, otherwise, assigns low value.
Then, we add the NCS values across the channel axis to get a 2D spatial heatmap.
We highlight that, unlike conventional visualization tools, our SAM does not require target class label to find a contribution or visual explanation \cite{zhou2016learning,selvaraju2017grad}.

Further, by using SAM, we investigate various configurations of SNNs. Firstly, we compare the internal spiking behavior of two different SNN training methods: surrogate gradient  based training \cite{kim2020revisiting} and ANN-SNN \cite{sengupta2019going} conversion on a non-trivial image dataset (\ie, Tiny-ImageNet).
Then, we observe the spike representation of each layer across different time-steps to understand the temporal characteristics of SNNs. We also analyze the effect of varying factors such as a leak rate and related hyperparameters on SAM and overall prediction. 
Finally, we provide a visual understanding of previously observed results \cite{sharmin2020inherent} that SNNs are more robust to adversarial attacks \cite{goodfellow2014explaining}. We measure the difference of heat maps between clean samples and adversarial samples using SAM to highlight the robustness of SNNs with respect to ANNs.

In summary, our key contributions can be summarized  as follows: 
(i) For the first time, we introduce a novel visualization technique for SNNs, called Spike Activation Map (SAM). We circumvent the non-differentiability problem of LIF neuron by calculating an attention map based on short ISI spikes in neurons. 
(ii) Interestingly, we find that SAM shows reliable visualization results without any ground truth class labels. 
{
(iii) By using SAM, we visualize and analyze the temporal characteristics and internal spike behavior of SNNs across various configurations, such as, training schemes, temporal parameters, adversarial inputs. Overall, our proposed SAM opens up the possibility towards interpretable and reliable neuromorphic computing.
}

\vspace{-1mm}

\section{Related Work}
\vspace{-1.5mm}

\subsection{Spiking Neural Networks}
\vspace{-1mm}

Spiking   Neural   Networks   (SNNs)   have   recently emerged as the next generation AI due to their huge energy efficiency benefits on asynchronous neuromorphic hardware.
Following the recent development of neuromorphic computing architectures such as TrueNorth \cite{akopyan2015truenorth} and Loihi \cite{davies2018loihi}, training algorithm of SNNs has received huge attention. 
% STDP
One intriguing learning algorithm is spike-timing-dependent  plasticity  (STDP)  \cite{bi1998synaptic} with a bio-plausible  Hebbian  learning  rule \cite{hebb2005organization}.
 This algorithm is based on local learning by using the spike correlation of pre-synaptic spikes and post-synaptic spikes. So far, STDP-based learning has been confined to shallow networks on small-scale datasets due to the absence of a global optimization rule.
% Conversion
Another widely-used method is ANN-SNN conversion method \cite{sengupta2019going,han2020rmp,diehl2015fast,rueckauer2017conversion}, which converts a pre-trained ANN to an SNN.
%Such methods aim to transfer ReLU activation to IF spiking activity without leak behavior. 
Since networks are trained in ANN domain, the training complexity is significantly removed. 
With careful threshold (or weight) balancing \cite{diehl2015fast}, ANN-SNN conversion shows good performance on large-scale datasets. It is worth mentioning that temporal dynamics are not considered in the process of training for converted SNNs.
%Unfortunately, the converted SNN requires a large latency for inference to emulate the floating point activation in ANN. 
% Moreover, conversion cannot exploit the leak behavior that is a biological hallmark of neurons in the human brain. 
% Surrogate
Recently, training SNNs with backpropagation \cite{neftci2019surrogate,lee2020enabling,wu2018spatio,kim2020revisiting} has been studied because it can take into account temporal neuronal dynamics during surrogate gradient descent. %Hence, the number of time-steps for both training and inference is significantly lesser than the conversion method.  
% However, the discrepancy between real gradient function and approximated gradient limit the depth of SNNs into shallow layers.
Despite the huge progress in training methods for SNNs, there is little attention given to the internal spike behavior of SNNs. Therefore, in this paper, we focus on SNN interpretability. Our results show that surrogate methods which have explicit temporal dependence during training are more interpretable than conversion.

\vspace{-1mm}

\subsection{Visualization Tools for ANNs}
\vspace{-1mm}
The interpretation of prediction in neural networks has received considerable attention due to its practicality in real-world scenarios.
Class Activation Map (CAM) \cite{zhou2016learning} highlights the  discriminative region of an image by using a global average pooling layer at the end of the feature extractor.
The CAM heat map is obtained by summing the feature maps at the last convolutional layer.
Several variations of CAM have been proposed
\cite{yang2020combinational,wang2020score,shi2020zoom}.
However, the necessity of the global average pooling layer in CAM limits its usage.
To address this issue, Selvaraju \etal proposed Grad-CAM \cite{selvaraju2017grad}, which is the generalized version of CAM.
Grad-CAM computes backward gradients from the classifier to a given intermediate layer where visual explanation is required.
Thus, the contribution of each neuron to the classification result can be quantified with the corresponding gradient value.
Then, a 2D heatmap is obtained by using the weighted sum of the activations across the channel axis based on the gradient value. In this work, we justify that directly applying Grad-CAM to calculate visual explanations in SNNs does not yield accurate results due to the non-differentiable nature of LIF neuron as well as non-dependence on temporal dynamics.

% \subsection{Adversarial Attack}

% Previous studies show that deep neural networks is vulnerable to adversarial inputs.
% Adversarial patch methods add a small patch on an image. This patch induces adversarial effect to the networks.
% However, Subramanya \etal \cite{subramanya2019fooling} assert that these methods are easily detected by Grad-CAM, which limits its practicality.
% Another way to generate adversarial attack is adding imperceptible noise to an input image.
% A FGSM attack \cite{goodfellow2014explaining} is widely-used and fundamental attack method.
% They compute the sign of gradient in direction of reducing the confidence of original prediction.
% Recently, Sharmin \etal \cite{sharmin2020inherent} propose a SNN-crafted FGSM attack. They accumulate gradients across whole time-step.
% We find that Grad-CAM shows a similar results on clean and adversarial examples. In contrast, SNNs show the potential to detect adversarial examples using its temporal dynamic.

% \clearpage

\section{Background}
\label{sec: background}
\vspace{-1mm}

% \subsection{Poisson Rate Coding}
% \vspace{-1mm}

\textbf{Poisson Rate Coding}: To convert a static image into multiple binary spikes, we use Poisson rate coding, or rate-based coding.
This is based on the human visual system \cite{adrian1926impulses}, and shows outstanding performance among various spike coding schemes such as temporal \cite{mostafa2017supervised}, phase \cite{kim2018deep}, and burst \cite{park2019fast}.
Poisson coding generates a spike train over multiple time-steps where the number of spikes is approximately proportional to the pixel intensity of the input image.
In practice, we compare each pixel value with a random number $[0, 255]$ at every time-step.
If the generated random number is less than the pixel intensity, the Poisson spike generator does not produce spikes, otherwise, it generates a spike with amplitude $1$. The generated spikes are then passed through an SNN. %the spiking networks across multiple time-steps.

% \subsection{Leaky-Integrate-and-Fire Neuron}
% \label{sec:method_LIFneuron}
% \vspace{-1mm}

\textbf{Leaky-Integrate-and-Fire Neuron}: Leaky-Integrate-and-Fire (LIF) neuron is the main  component of SNNs.
The internal state of an LIF neuron is represented by a membrane potential $U_m$.  
As time goes on, the membrane potential decays with time constant  $\tau_m$.
Given an input signal $I(t)$ and an input register $R$ at time $t$, the differential equation of the LIF neuron can be formulated as:
\begin{equation}
    \tau_m \frac{dU_m}{dt} = -U_m  + RI(t).
    \label{eq:LIF_origin}
\end{equation}

This continuous dynamic equation is converted into a discrete equation for digital simulation. More concretely, we formulate the membrane potential $u_{i}^{t}$ of a single neuron $i$ as:
\begin{equation}
    u_i^t = \lambda u_i^{t-1} + \sum_j w_{ij}o^t_j - \theta o^{t-1}_i,
    \label{eq:LIF}
\end{equation}
where, $\lambda$ is a leak factor, $w_{ij}$ is  the weight of the connection between pre-synaptic neuron $j$ and post-synaptic neuron $i$.
If the membrane potential $u_i^{t-1}$ exceeds a firing threshold $\theta$, the neuron $i$ generate spikes $o_i^{t-1}$, which can be formulated as:
\vspace{-1.5mm}
\begin{equation}
    o^{t-1}_i =
\begin{cases}
 1,          & \text{if $u_i^{t-1}>\theta$},  \\
    0
    & \text{otherwise.} 
\end{cases}
\label{eq:firing}
\end{equation}
% \vspace{-1.5mm}
After the neuron fires, we perform a soft reset, where the membrane potential value is lowered by threshold $\theta$. 
Because of this non-differentiable firing behavior,
training SNNs with gradient learning is a huge challenge \cite{roy2019towards}.
To address this issue, previous studies \cite{neftci2019surrogate,lee2020enabling} approximate the backward gradient function (\eg, piecewise linear and exponential) to implement gradient learning.
Fig. \ref{fig:neuron_dynamic} illustrates the membrane potential dynamics of an LIF neuron.

\section{Methodology}
\vspace{-1mm}
% In this section, we present the details of two representative training methods of SNNs (\ie surrogate gradient training and conversion) that we use for our case study.
% Then we explain SNN-crafted Grad-CAM which accumulates the backward gradient across time-steps.
% Finally, we propose SAM, a new concept of visualization for SNNs.

% \subsection{Training Deep Spiking Neural Networks}
% \vspace{-1mm}

In this paper, we visualize the internal spike behavior of 
two representative and widely-used training methods: surrogate gradient training \cite{kim2020revisiting} and ANN-SNN conversion \cite{sengupta2019going}.
Since ANNs can be trained with well-established optimization methods and frameworks, SNNs from ANN-SNN conversion shows reliable performance on very large-scale datasets (\eg, ImageNet).
In contrast, most surrogate gradient training methods are limited to small datasets (\eg, MNIST and CIFAR10) due to approximated backward gradients.
These simple datasets are too small to be analyzed by visualizing heatmap.
% Therefore, to the best of our knowledge, the explainability of SNNs has not been explored with 2D heatmap visualization.
But, the authors in \cite{kim2020revisiting} recently proposed temporal adaptive batch normalization (BN) for surrogate gradient learning, enabling training on larger datasets such as CIFAR100 and Tiny-ImageNet.
We exploit this algorithm for the case study of surrogate gradient training to compare with ANN-SNN conversion on Tiny-ImageNet dataset.

\subsection{Surrogate Gradient Backpropagation}

The SNN-crafted BN layer \cite{kim2020revisiting}, called Batch Normalization Through Time (BNTT), improves training stability and reduces latency while preserving  classification accuracy.
We add the BNTT layer before an LIF neuron.
Therefore, the weighted pre-synaptic input spikes are normalized as: 
%
% \vspace{-1mm}
\begin{equation}
    u_i^t    = \lambda u_i^{t-1} + \gamma_i^t (\frac{\sum_j w_{ij}o^t_j  - \mu^t_i}{\sqrt{(\sigma_i^t)^2 + \epsilon}}) - \theta o^{t-1}_i,
    \label{eq:LIFwithBN}
\end{equation}
% \vspace{-1mm}
where, $\gamma_i^t$ is a learnable parameter in the BNTT layer, $\epsilon$ is a small constant for numerical  stability, the mean $\mu_{i}^t$ and variance $\sigma_{i}^t$ are calculated from the samples in a mini-batch for each time step $t$.  We append all intermediate layers of an SNN with a BNTT layer.
At the output layer, we set the number of output neurons  to the number of classes $C$.
To prevent information loss from the leakage of a neuron, we accumulate the spikes over all time-steps by fixing the leak parameter $\lambda$  (Eq. \ref{eq:LIF}) as one. 
This stacked voltage is converted into probability distribution using a softmax layer. 
Finally, we compute the cross-entropy loss as: 
\vspace{-2mm}
\begin{equation}
    {L} = - \sum_{i} y_{i} log(\frac{e^{u_i^T}}{\sum_{k=1}^{C}e^{u_k^T}}).
    \label{eq:celoss}
    \vspace{-2mm}
\end{equation}
Here, $y_i$ represents the ground truth label, and $T$ is the total number of time-steps.

% \begin{algorithm}[t]\small
%         \caption{Surrogate Gradient Backpropagation}
%     %   \hspace*{\algorithmicindent} 
%       \textbf{Input}: Mini-batch ($X$); label set ($Y$); max\_timestep ($T$)\\
%       \textbf{Output}:  Updated network weights 
%       \begin{algorithmic}[1]
%         % \State{\textbf{begin}}
%         %
%         \For{$i \gets 1$ to $max\_iter$}
%             \State {// fetch a mini batch X}
%             \For{$t \gets 1$ to $T$}  
%                 \State{O $\leftarrow$ PoissonGenerator(X)}
%                 %
%                 \For{$l \gets 1$ to $L-1$} 
%                     \State{$(O^t_{l}, U_l^{t}) \leftarrow (\lambda, U_l^{t-1}, \textup{BNTT}_{t}(W_{l}, O^{t-1}_{l-1}))$}
%                 \EndFor
%                 \State{// for final layer, stack the voltage}
%                 \State{$U_L^{t} \hspace{-1mm} \leftarrow  \hspace{-1mm} ( U_L^{t-1}, \textup{BNTT}_{t}(W_{L}, O^{t-1}_{L-1}))$ }
%             \EndFor
%             \State{// Calculate the loss and back-propagation}
%             \State{$L \leftarrow (U_L^T, Y)$}
%         \EndFor
%         % \State{\% Early exit}
%         % \State{$t_{exit} \leftarrow (\gamma^t)$}
%       \end{algorithmic}
%           \label{algorithm: overall}
% \end{algorithm}

% \textbf{Backpropagation:} 
%
%A backpropagation process is especially important in SNNs since the functionality of an LIF neuron is non-differentiable.
Then, we accumulate the backward gradients over all time-steps, which is called back-propagation through time (BPTT)  \cite{neftci2019surrogate}.
The accumulated gradients at hidden layers and the output layer can be represented as:  
\vspace{-1mm}
\begin{equation}
    \Delta W_l = \sum_{t} \frac{\partial L}{\partial W_l^t} =
\begin{cases}
  \sum_{t}\frac{\partial L}{\partial O_l^t}\frac{\partial O_l^t}{\partial U_l^t}\frac{\partial  U_l^t}{\partial W_l^t},              & \text{$l$: hidden layer} \\
    \sum_{t}\frac{\partial L}{\partial U_l^T}\frac{\partial U_l^T}{\partial W_l^t},
    & \text{$l$: output layer} 
\end{cases}
\label{eq:delta_W}
\vspace{-1mm}
\end{equation}
where, $W_l$, $O_l$ and $U_l$ stand for weight matrix, output spike matrix, and membrane potential matrix at layer $l$, respectively. 
As the output layer does not generate spikes, we compute the exact derivative of the loss $L$ with respect to the membrane potential $u_i^T$:
\vspace{-2mm}
\begin{equation}
\frac{\partial L}{\partial u_i^T} =\frac{e^{u_i^T}}{\sum_{k=1}^{C}e^{u_k^T}} - y_i.
\vspace{-2mm}
\end{equation}
However, for hidden layer, the gradient term $\frac{\partial o_i^t}{\partial u_i^t}$ is not differentiable due to the firing behavior of an LIF neuron.
Therefore, $\frac{\partial o_i^t}{\partial u_i^t}$ should be formulated with an approximated continuous function (Fig. \ref{fig:neuron_dynamic}).
To this end, we use a piecewise linear function:
\vspace{-1mm}
\begin{equation}
    \frac{\partial o_i^t}{\partial u_i^t} = \beta \max \{0, 1-  \ | \frac{u_i^t - \theta}{\theta} \ | \},
    \label{eq:approximated_gradient}
\vspace{-1mm}
\end{equation}
where, $\beta$ is a scaling factor for the gradient value. We set $\beta$ as 0.3 to prevent a gradient exploding problem.
% Note that $x_i^t = \sum_j w_{ij}o^t_j  $ in Eq. \ref{eq:delta_W} is an input signal to the BNTT layer, which is differentiable with respect to the weight matrix $W_l^t$.
Based on the gradient value, the weights of SNNs are updated. 
% Algorithm 1 summarizes the whole training process of surrogate gradient training.

\begin{algorithm}[t]\small
        \caption{ANN-SNN Conversion}
    %   \hspace*{\algorithmicindent} 
       \textbf{Input}: Input set ($X$); label set ($Y$); max\_timestep ($T$); pre-trained ANN model (A); SNN model (S); total layer number (L)   \\
      \textbf{Output}:  Updated SNN network with threshold balancing
      \begin{algorithmic}[1]
        % \State{\textbf{begin}}
        %
        \State {// Copy ANN weights to SNN}
        \State {$S.weights \leftarrow A.weights$}
        \State {// Initialize threshold voltage}
        \State {$S.th$ $\leftarrow$ 0}
        \For{$l \gets 1$ to $L-1$}
            \For{$t \gets 1$ to $T$}  
                \State{$O^{t}$ $\leftarrow$ PoissonGenerator(X)}
                \For{$l_{tmp} \gets 1$ to $l$} 
                    \If {$l_{tmp} < l$}
                        % \State{\textup{// Forward propagation}}
                        \State{$(O^t_{l}, U_l^{t}) \leftarrow 
                        ( U_l^{t-1}, W_{l}, O^{t-1}_{l-1})$}
                    % \EndIf
                    \Else
                        % \State{Activation $\leftarrow$ ${S_{l}(O^t_{l-1})}$}
                        \State{\textup{// Threshold update for each layer}}
                        \State{$S_{l}.th$ $\leftarrow$ max($S_{l}.th$, ${W_{l}O^{t-1}_{l-1}}$)}
                    \EndIf
                \EndFor
            \EndFor
        \EndFor
        % \State{\% Early exit}
        % \State{$t_{exit} \leftarrow (\gamma^t)$}
      \end{algorithmic}
          \label{algorithm: overall}
\end{algorithm}
% \vspace{-3mm}

% \begin{figure}[t!]
% \begin{center}
% \def\arraystretch{0}
% \vspace*{-0.03in}
% \begin{tabular}{@{\hskip 0.01\linewidth}c@{\hskip 0.07\linewidth}c@{\hskip 0.07\linewidth}c}
% \includegraphics[width=0.45\linewidth]{figures/layer2.png} &
% \includegraphics[width=0.45\linewidth]{figures/layer4.png} 
% % &\includegraphics[width=0.275\linewidth]{figures/layer6.png}
% \vspace{2mm}
% \\
% { (a) Conv2} & {(b) Conv4 }\\
% % & {\hspace{4mm}(c) Conv6 }\\
% \end{tabular}
% % \vspace*{-0.1in}
% \end{center}
% \caption
% {
% Visualization of the SNN-crafted Grad-CAM at (a) Conv2, (b) Conv4, and  (c) Conv6 in a VGG11-like SNN architecture trained on TinyImageNet dataset. Each layer requires one time-step to propagate spike to the next layer. Therefore, heatmaps cannot be achievable before $t=11$ (\ie, the depth of VGG11).  The  SNN-crafted Grad-CAM becomes non-discriminative in shallow layers due to the approximated backward gradients.
% }
% \vspace{-3mm}
% \label{fig:Srad_CAM_problem}
% \end{figure}

\subsection{ANN-SNN Conversion}

We use \cite{sengupta2019going} for implementing the ANN-SNN conversion method.
They normalize the weights or the firing threshold ($\theta$ in Eq. \ref{eq:firing}) to take into account the actual SNN operation in the conversion process.
The overall algorithm for the conversion method is shown in Algorithm 1.
First, we copy the weight parameters of a pre-trained ANN to an SNN.
Then, for every layer, we compute the maximum activation across all time-steps and set the firing threshold to the maximum activation value.
The conversion process starts from the first layer and sequentially goes through deeper layers.
Note that we do not use BN \cite{ioffe2015batch} since all input spikes have zero mean values.%, resulting in non-discriminative firing threshold values.
Also, following the previous works \cite{han2020rmp,sengupta2019going,diehl2015fast}, we use Dropout \cite{srivastava2014dropout} for both ANNs and SNNs.

\subsection{SNN-crafted Grad-CAM}

Grad-CAM \cite{selvaraju2017grad} highlights the region of the image that highly contributes to classification results.
Grad-CAM computes a backward gradient from the classifier logit to the pre-defined target layer.
After that, channel-wise attention value is obtained by using global average pooling.
Based on this, the final heatmap is defined as the weighted sum of feature maps.
Different from conventional deep neural networks, SNNs take spike trains across multiple time-steps.
Therefore, we can compute multiple SNN-crafted Grad-CAMs across the total number of time-steps $T$.
Similar to Grad-CAM, we quantify the contribution of each channel by accumulating gradients across all time-steps:
\begin{equation}
    \alpha^{c, k} = \frac{1}{N}\sum _i\sum _j\sum _t \frac{\partial y^c}{\partial A^k_{ij, t}}.
\end{equation}
Here, $N$ is a normalization factor, and 
$A^k_{ij, t}$ is the activation value of the $k$th channel at  time-step $t$, and (i, j) is the pixel location.
Note that we use a ground truth label $c$ for a given image to compute the heatmap.
Therefore, the channel-wise weighted sum of spike activation can be calculated as:
\begin{equation}
    G^c_{ij,t} = max(0, \sum_{k}\alpha^{c,k}_{t}A^k_{ij,t}).
\end{equation}
For a clear comparison with conventional Grad-CAM, we call $G^c_{ij,t}$ as ``SNN-crafted Grad-CAM" afterward.

\begin{figure}
    \centering
    \includegraphics[width=0.38\textwidth]{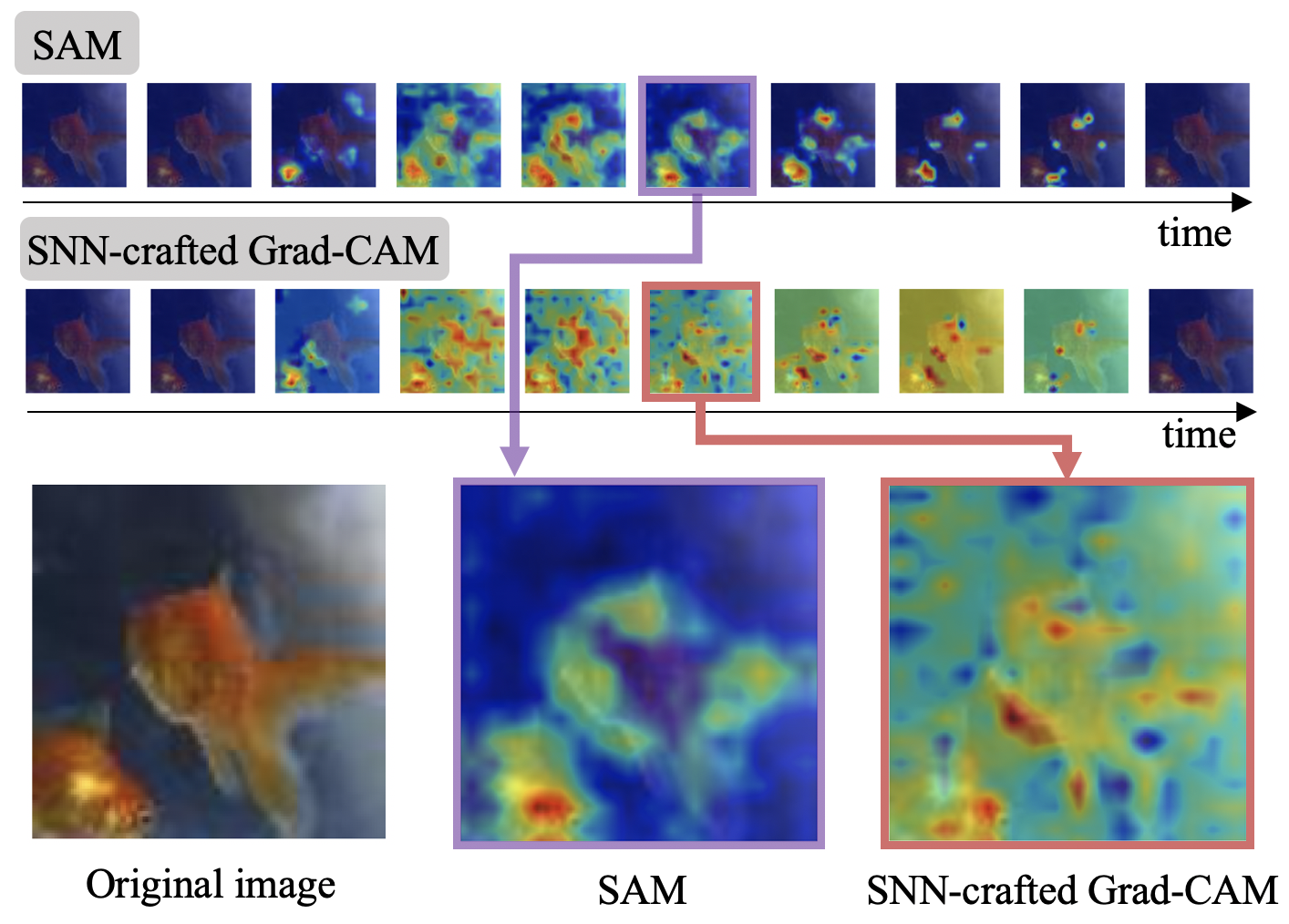}
    % \vspace{-3mm}
    \caption{ 
    Visualization of SNN-crafted Grad-CAM and SAM at  Conv4 in VGG11 on Tiny-ImageNet dataset. 
    We use surrogate gradient training (the conversion method also shows similar results). The approximated backward gradient function in SNN-crafted Grad-CAM induces ``heatmap smoothing effect". In contrast, the proposed SAM visualization highlights the discriminative region of the image. 
    }
    \label{fig:heatmap_smoothing_effect}
    \vspace{-4mm}
\end{figure}

\begin{figure}[t!]
\begin{center}
\def\arraystretch{0}
\vspace*{-0.03in}
\begin{tabular}{@{\hskip 0.07\linewidth}c@{\hskip 0.07\linewidth}c@{\hskip 0.07\linewidth}c}
\includegraphics[width=0.40\linewidth]{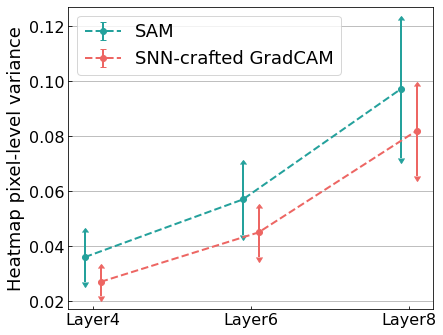} &
\includegraphics[width=0.40\linewidth]{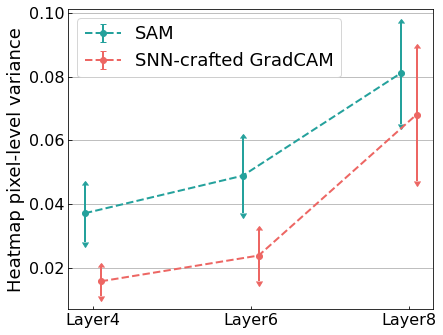} 
% &\includegraphics[width=0.275\linewidth]{figures/layer6.png}
\vspace{2mm}
\\
{ (a) Surrogate Gradient} & {(b) Conversion }\\
% & {\hspace{4mm}(c) Conv6 }\\
\end{tabular}
\vspace{-2mm}
\end{center}
\caption
{ Pixel-level variance in heatmaps obtained from surrogate gradient learning and conversion. We report the average variance from the total samples in Tiny-ImageNet. 
For all scenarios, SAM shows a higher hearmap variance compared to SNN-crafted Grad-CAM that suffers from the approximated backward gradient.
}
\vspace{-3mm}
\label{fig:gradient_error_problem}
\end{figure}

However, SNN-crafted Grad-CAM suffers from what we term as a ``heatmap smoothing effect" caused by the approximated backward gradient function.
To visualize the heatmap at shallow/initial layers, the gradients need to pass through multiple layers using the approximated backward function (Eq. \ref{eq:approximated_gradient}).
The accumulated approximation error induces non-discriminative heatmap as shown in Fig. \ref{fig:heatmap_smoothing_effect}.
Note that the beginning and end of time-steps have few spike activity \cite{kim2020revisiting} resulting in heatmaps with zero values (see Fig. \ref{fig:heatmap_smoothing_effect}). 
To validate the ``heatmap smoothing effect" quantitatively, we compute the pixel-wise variance of the heatmap. Thus, the heatmap containing non-discriminative information (\ie similar pixel values) should have lower variance.
In Fig. \ref{fig:gradient_error_problem},  SNN-crafted Grad-CAM shows lower variance compared to our proposed SAM (will be discussed in the next section).
Note that there are multiple heatmaps (one heatmap per time-step) in SNN visualization. So, we use the maximum variance value across all time-steps in Fig. \ref{fig:gradient_error_problem}. {Further, we note that the visualization in both SAM and SNN-crafted Grad-CAM in Fig. \ref{fig:heatmap_smoothing_effect} varies across each time-step underlying the fact that the SNN looks at different regions of the same input over time to make a prediction.}
Overall, the visualization tool for SNNs requires a new perspective that can circumvent the error accumulation problem in backpropagation.

\subsection{Spike Activation Map (SAM)}

% The objective of Spike Activation Map (SAM) is to circumvent the error accumulation problem of the SNN-crafted Grad-CAM and achieves meaningful heat map visualization.
% To this end, we exploit the physiological evidence.
% Previous studies \cite{reich2000interspike,snider1998burst,shih2011improved} have witnessed that spikes having short inter-spike interval (ISI) contribute to the neural decision process more than the other spikes.
% This is because short-ISI spikes are more likely to stimulate a postsynaptic neurons, resulting in a larger number of post-spikes \cite{alonso1996precisely,lisman1997bursts,snider1998burst}.
% Fig. \ref{fig:ISI} illustrates the definition of ISI, the time difference between successive spikes.

%  for each neuron, wecompute aneuronal contribution score(NCS) regarding theclassification result. The NCS score is defined as the sum oftemporal spike contribution score(TSCS) with exponentialkernel. The TSCS score assigns the high value for spikes fir-ing within a short time window, otherwise, assigns the lowvalue.  Then we adds the NCS across channel axis to get a2D spatial heatmap at every time-step.

We present a new paradigm for the visualization of SNNs.
We do not use any class label for backpropagation, and only use the spike activity in forward propagation.
Thus, this heatmap is not just for a specific class but highlights the regions that the network focuses for any given image. %We note that SAM is a general and unsupervised visual explanation tool.
Surprisingly, we observe that SAM shows meaningful visualization even without any ground truth labels (see Section \ref{sec: Unsupervised Visualization Tool}).
Mathematically, our objective can be formulated as to find a mapping function $f(\cdot)$:
\begin{equation}
    M_{t} \leftarrow f(S_{0}, S_{1}, ..., S_{t-1}),
    \label{eq:definition_sam}
\end{equation}
where, $M_{t}$ is SAM and $S_{t}$ is spike activity at time-step $t$. 
% In other words, we calculate the spike activation map based on previous spike activities without the need of backpropagation.
%
%
In this paper, we use the biological observation that spikes having short inter-spike interval (ISI) highly contribute to the neural decision process \cite{reich2000interspike,snider1998burst,shih2011improved}.
This is because short-ISI spikes are more likely to stimulate  post-synaptic neurons, conveying more information \cite{alonso1996precisely,lisman1997bursts,snider1998burst}.
To apply this to our visualization method, we first define the temporal spike contribution score (TSCS) which evaluates the contribution of previous spike at time $t'$ to the current time $t$ in the same neuron. The TSCS value can be formulated as:
\begin{equation}
    T(t, t') = \exp(- \gamma |t-t'|),
    \label{eq:tcsc}
\end{equation}
where, $\gamma$ is a hyperparameter which controls the steepness of the exponential kernel function.

\begin{figure*}[t!]
%   \vspace{-20pt}
  \begin{center}
    \includegraphics[width=0.75\textwidth]{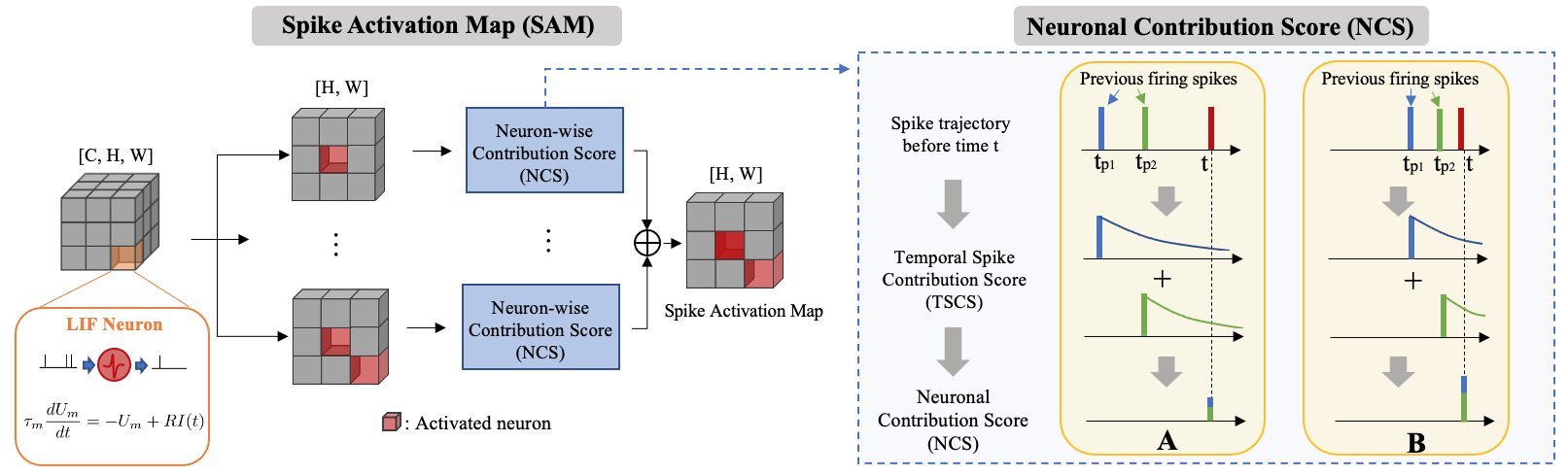}
  \end{center}
  \vspace{-3mm}
  \caption{Illustration of spike activation map (SAM). For each channel, we compute a neuron-wise contribution score. After that, we sum all neuronal contribution score (NCS) map in the
  channel axis. 
  The NCS for each neuron is based on the previous spike trajectory. 
  For every spike, we define temporal spike contribution score (TSCS) with an exponential kernel.
  We take into account TSCS from previous spikes in order to compute NCS.}
  \vspace{-4mm}
   \label{fig:SAM}
\end{figure*}

To consider multiple previous spikes, we define a set $P_{ij}^k$ that consists of previous firing times of a neuron at location $(i,j)$ in $k$th channel.
For every time-step, we compute a neuronal contribution score (NCS)  $N^k_{ij,t}$  at time-step $t$, by adding all TSCS of spikes in $P_{ij}^k$:
\begin{equation}
    N^k_{ij,t} = \sum_{t' \in P_{ij}^k} T(t, t').
    \label{eq:ncs}
\end{equation}
% where, $\gamma$ is hyperparameters in which controls the steepness of the kernel function, $t'$ is the element of previous firing time set $P$, and $(i, j)$ is a pixel location. 
% Thus, the value is determined according to the difference between current time and previous spike trajectories.
Thus, a neuron has a high NCS if large number of spikes are fired in a short time interval and vice-versa.
Finally, we calculate the SAM heatmap $M_{ij,t}$ at time-step $t$ and location $(i, j)$ by multiplying spike activity $S_{ij,t}$ with NCS value $N_{ij,t}$:
\begin{equation}
    M_{ij,t} = \sum_{k} N_{ij,t}^k  S^k_{ij,t}.
    \label{eq:sam}
\end{equation}

We illustrate the overall flow of SAM in Fig. \ref{fig:SAM}. 
For every neuron, we compute NCS and add the values across the channel axis in order to get SAM.
To elaborate, we depict two examples (case A and case B) for calculating NCS.
In case $A$, the previous spikes fire at time-step $t_{p1}$ and $t_{p2}$, a long time before the current spike time $t$.
As a result, the contribution of previous spikes is small due to the exponential kernel.
On the other hand, in case $B$, $t_{p1}$ and $t_{p2}$ are close to the current spike time $t$. In this case, the neuron has a high NCS value.

\textbf{Discussion:} Overall, without any requirement for backpropagation and ground truth labels, we can visualize the discriminative region by using the concept of inter-spike interval.  We would like to note that SAM cannot be applied to ANNs due to the real-valued static nature of input processing in ANNs. So far, SNNs have been explored as an energy-efficient alternative to ANNs. With SAM, for the first time, we bring out the interpretability advantage of the temporal dynamics in SNNs over static ANNs.  %That is, varying spike information over time in SNNs can be used for interpretability.
 We assert that the proposed SAM is hardware friendly since all computations are in forward propagation. 
Therefore, our SAM can be used as a practical interpretation tool for future neuromorphic computing applications.

\section{Experiments}

% In  this  section,  we  carry  out  comprehensive  experiments with SAM.  We first scrutinize the internal behavior of surrogate gradient and ANN-SNN conversion.
% We then present ablation  studies  of  the  main  components of SNNs quantitatively and qualitatively. Finally, we justify the robustness of SNNs against adversarial attack using SAM.

\begin{figure*}[t!]
%   \vspace{-20pt}
  \begin{center}
    \includegraphics[width=0.80\textwidth]{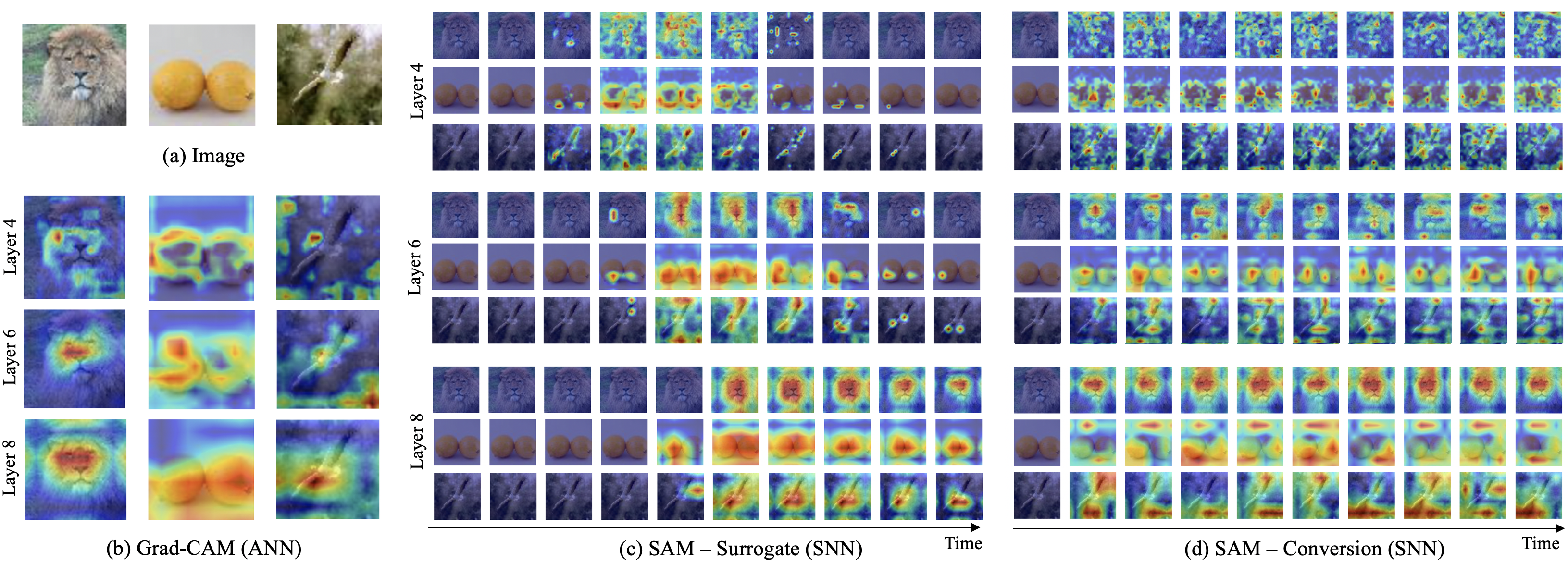}
  \end{center}
  \vspace{-4mm}
  \caption{Visualization of the internal spike representation of VGG11 using SAM at layer 4, layer 6, and layer 8. We show the visualization for 10 uniformly sampled time-steps.
  It is worth mentioning Grad-CAM  exploits  ground truth labels but  our  SAM  can be obtained without any label information. 
%   A shallow layer highlights low-level structure. In contrast, deep layers visualize semantic representation. BNTT shows better localization compared to conversion. 
%   We use a VGG11 architecture and uniformly sample 10 images since SNNs have multiple time-steps.
  }
  \vspace{-5mm}
   \label{fig:sam_grad_visualization}
\end{figure*}

\vspace{-1mm}

\subsection{Experimental Setup}
\vspace{-1mm}
\textbf{Dataset and Network:} To conduct a comprehensive analysis, we carefully select the dataset for our experiments.
This is because smaller datasets such as MNIST, CIFAR10, and CIFAR100 have too low resolution (\eg, $28 \times 28$ or $32 \times 32$) to visualize.
ImageNet dataset has a high image resolution but directly training SNNs with surrogate gradient becomes hard. 
Therefore, we conduct a case study on the Tiny-ImageNet which is the modified subset of the original ImageNet dataset.
Tiny-ImageNet consists of 200 different classes of ImageNet dataset \cite{deng2009imagenet}, with 100,000 training and 10,000 validation images. The resolution of the images is $64 \times 64$ pixels.
Our implementation is based on Pytorch \cite{paszke2017automatic}.
We adopt a VGG11 architecture for both ANNs and SNNs.
For ANN-SNN conversion method, we use 500 time-steps  with firing threshold scaling \cite{han2020rmp}.  
For surrogate gradient training, we train the networks with standard SGD with momentum 0.9, weight decay 0.0005, time-steps 30.
The base learning rate is set to 0.1.
We use step-wise learning rate scheduling with a decay factor 10 at [0.5, 0.7, 0.9]  of the total number of epochs. 
We set the total number of epochs to 90.
We set the leak factor of SNN with  surrogate gradient learning and conversion to 0.99 and 1, respectively.
For visualization, we uniformly sample 10 images for both surrogate gradient learning and conversion.

\textbf{Evaluation Metric:} 
To quantitatively compare the SAM visualization of conversion and surrogate gradient, we use Grad-CAM obtained from ANNs as a reference.
% Note that we do not have to use class label for SAM.
To quantify the error between SAM and Grad-CAM, we compute the cross entropy function between the predicted SAMs (one SAM for one time-step) and a Grad-CAM from ANN at every time-step.
Then we select the minimum error across all time-steps and define the minimum value as a localization error.

% \begin{equation}
%     E = \min_{t} \{ \frac{1}{N} \sum_{i,j} G_{ij}\log(S_{ij,t}) + (1-G_{ij})\log(1-S_{ij,t}) \}.
%     \label{eq:localization_error}
% \end{equation}
% Here, $N$ is normalization factor and (i, j) indicates a pixel location.

% \begin{table}[t]
% \addtolength{\tabcolsep}{1.5pt}
% \centering
% \caption{Localization error comparison.}
% \resizebox{0.27\textwidth}{!}
% {
% \begin{tabular}{lcc}
% \toprule
%   & Surrogate & Conversion  \\
% \midrule
%     Layer 4 &  0.58  & 0.70  \\
%     Layer 6 & 0.72  & 0.74  \\
%     Layer 8  & 0.88 & 1.05   \\
% \bottomrule
% \end{tabular}%
% }
% \label{table: loc_comp}
% \end{table}
% %

% \begin{figure}
%     \centering
%     \includegraphics[width=0.27\textwidth]{figures/sam_localization_compare.png}
%     \vspace{1mm}
%     \caption{ Localization error comparison. We compare SAM from surrogate gradient learning and conversion with Grad-CAM.
%     }
%     \label{fig:loc_comp}
%     \vspace{-3mm}
% \end{figure}

\begin{figure}[t!]
\begin{center}
\def\arraystretch{0}
\vspace*{-0.03in}
\begin{tabular}{@{\hskip 0.01\linewidth}c@{\hskip 0.07\linewidth}c@{\hskip 0.07\linewidth}c}
\includegraphics[width=0.45\linewidth]{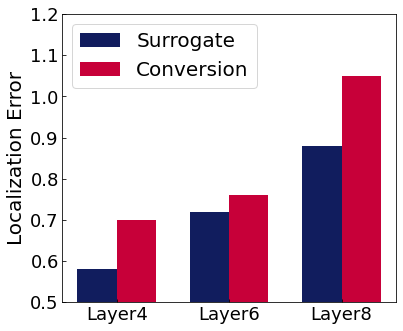} &
\includegraphics[width=0.43\linewidth]{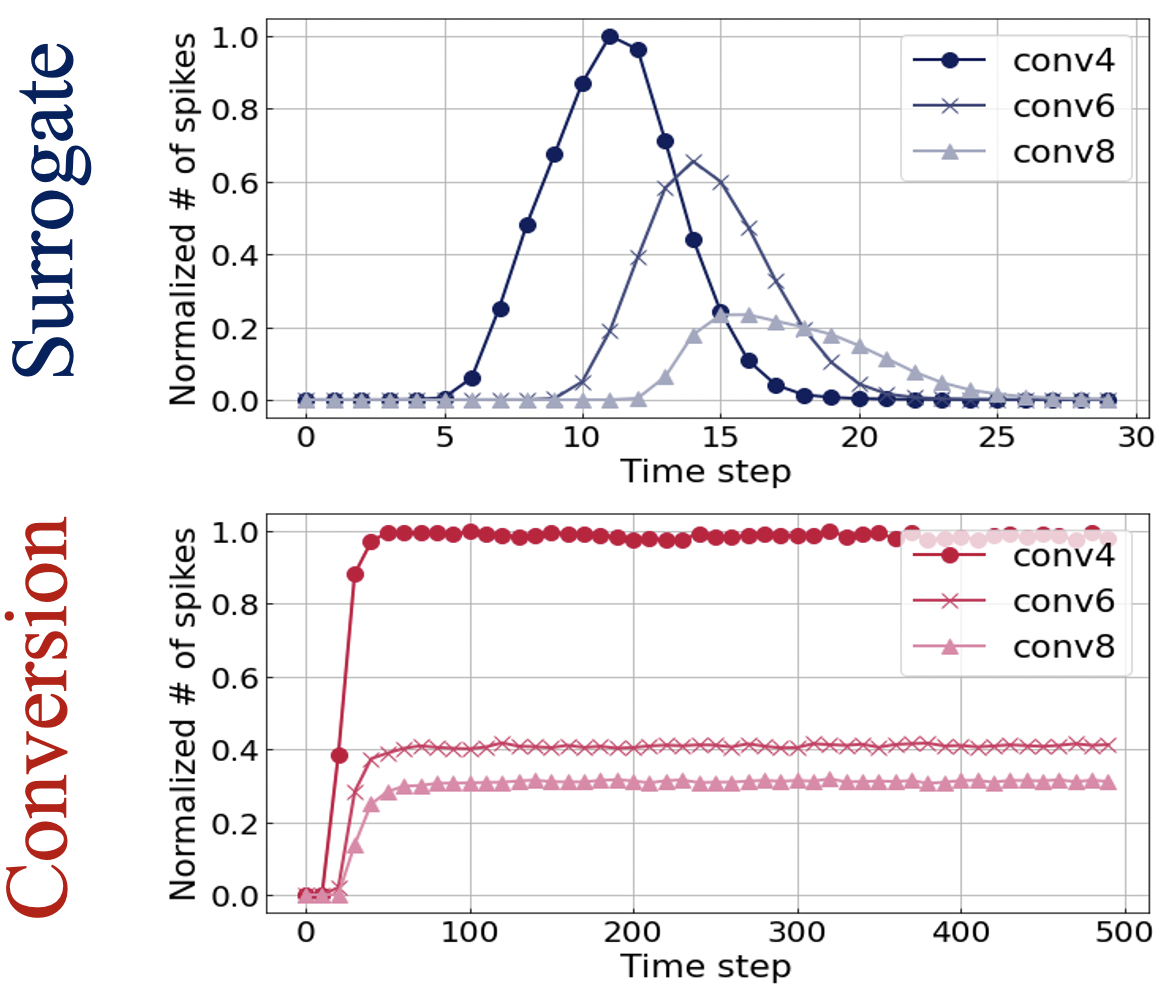} 
% &\includegraphics[width=0.275\linewidth]{figures/layer6.png}
\vspace{-0.2mm}
\\
{ (a) } & {(b) }\\
% & {\hspace{4mm}(c) Conv6 }\\
\end{tabular}
\vspace{-4mm}
\end{center}
\caption
{  (a) Localization error comparison. We compare SAM from surrogate gradient learning and conversion with Grad-CAM. (b) Visualization of the normalized number of spikes.
}
\vspace{-6mm}
\label{fig:loc_comp}
\end{figure}

% \begin{figure}[t]
% \begin{center}
% \def\arraystretch{0.5}
% \begin{tabular}{@{\hskip 0.015\linewidth}c@{\hskip 0.015\linewidth}c@{\hskip 0.015\linewidth}c}
% \includegraphics[width=0.5  \linewidth]{figures/leakfactor_local.png} &
% \includegraphics[width=0.5\linewidth]{figures/leakfactor_acc.png} 
% % \includegraphics[width=0.28\linewidth]{figures/early_tinyimagenet.png} 
% \\
% {\hspace{2.7mm} (a) } & {\hspace{2.7mm}(b)}\\
% \end{tabular}
% \end{center}
% \caption{Leak ablation}
% \label{fig:early_exit}
% \end{figure}

\begin{figure*}[t!]
%   \vspace{-20pt}
  \begin{center}
    \includegraphics[width=0.82\textwidth]{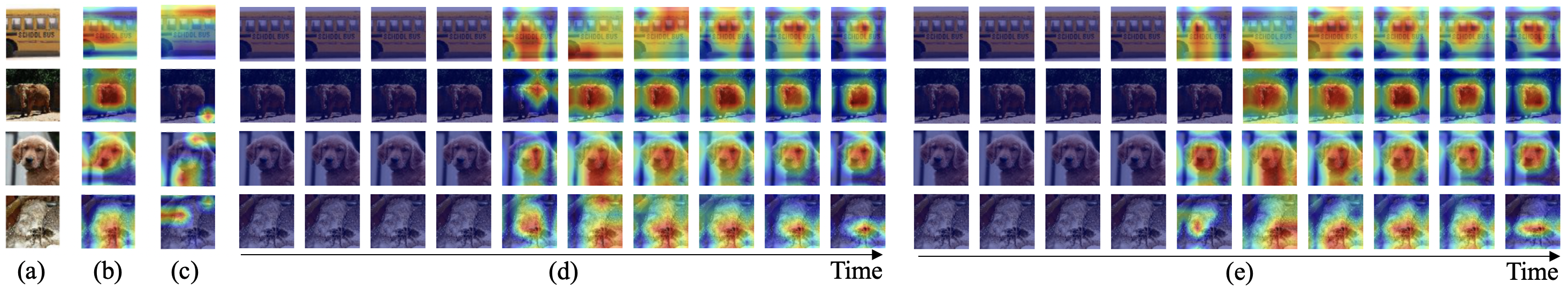}
  \end{center}
  \vspace{-5mm}
  \caption{ Visualization of robustness of SAM.
  (a) Original image. (b) Grad-CAM from ANN on clean input. (c) Grad-CAM from ANN on adversarial input. SAM from SNN trained with surrogate gradient learning on (d) clean inputs and
  (e) adversarial inputs.
  }
  \vspace{-4mm}
   \label{fig:sam_fgsm}
\end{figure*}

\begin{figure}[t!]
\begin{center}
\def\arraystretch{0}
\vspace*{-0.03in}
\begin{tabular}{@{\hskip 0.01\linewidth}c@{\hskip 0.07\linewidth}c@{\hskip 0.07\linewidth}c}
\includegraphics[width=0.43\linewidth]{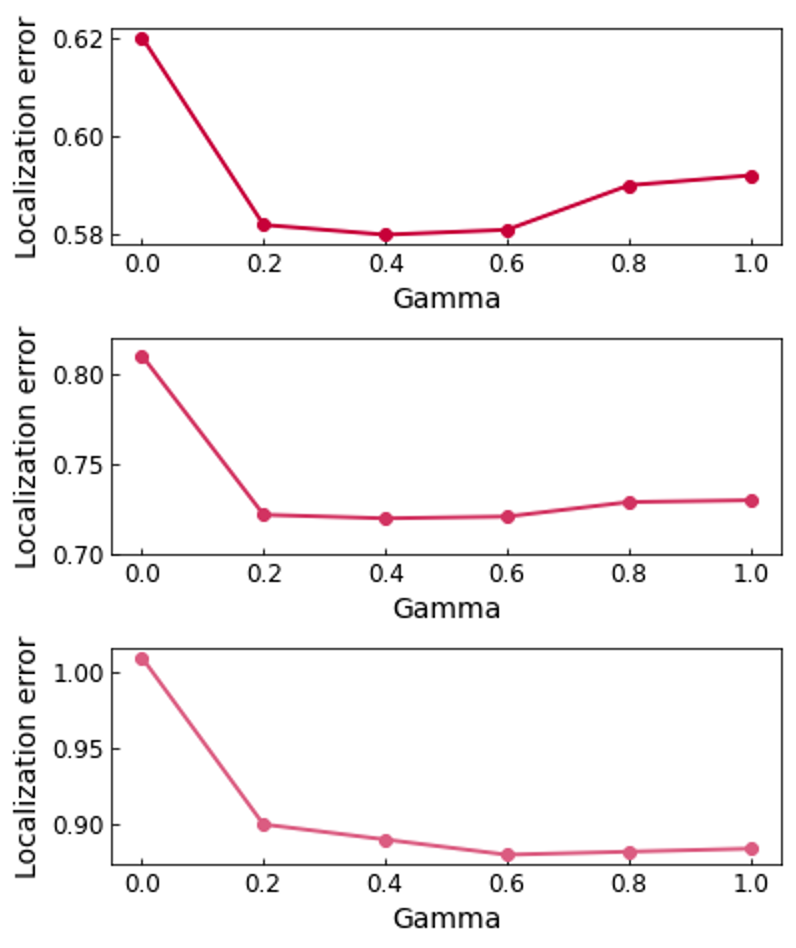} &
\includegraphics[width=0.43\linewidth]{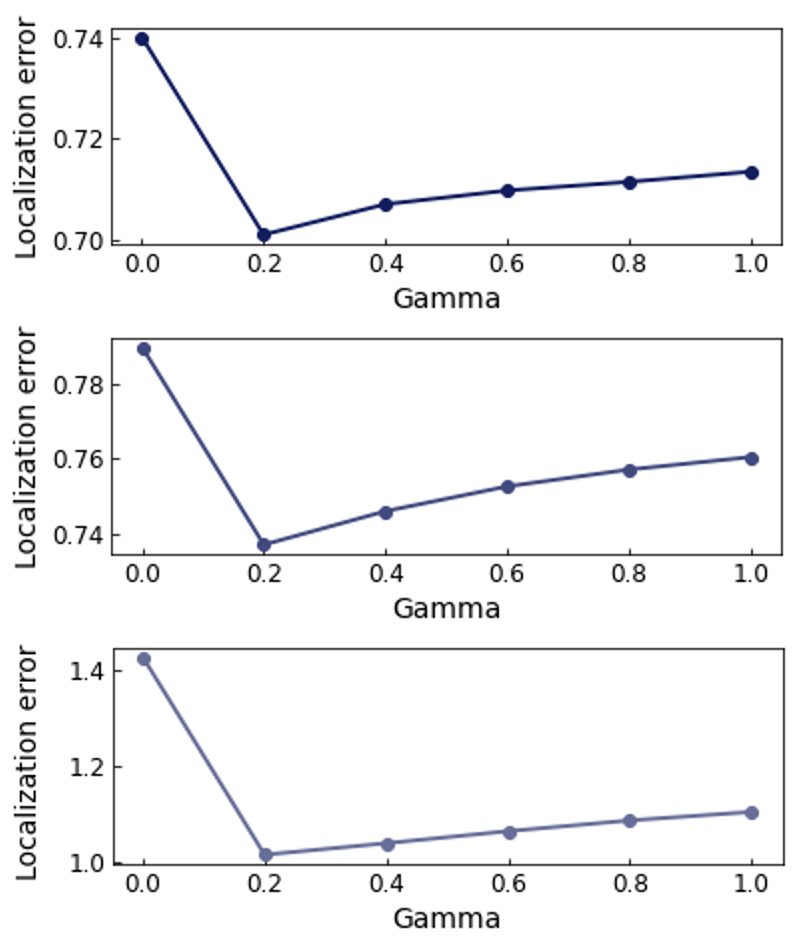} 
% &\includegraphics[width=0.275\linewidth]{figures/layer6.png}
\vspace{2mm}
\\
{ (a) Surrogate Gradient} & {(b) Conversion }\\
% & {\hspace{4mm}(c) Conv6 }\\
\end{tabular}
% \vspace*{-0.1in}
\end{center}
\caption
{  Localization error  at layer 4 (top row), layer 6 (middle row), and layer 8 (bottom row)  with respect to hyperparameter $\gamma$.
}
\vspace{-2.5mm}
\label{fig:gamma_ablation}
\end{figure}

\begin{figure}
    \centering
    \includegraphics[width=0.43\textwidth]{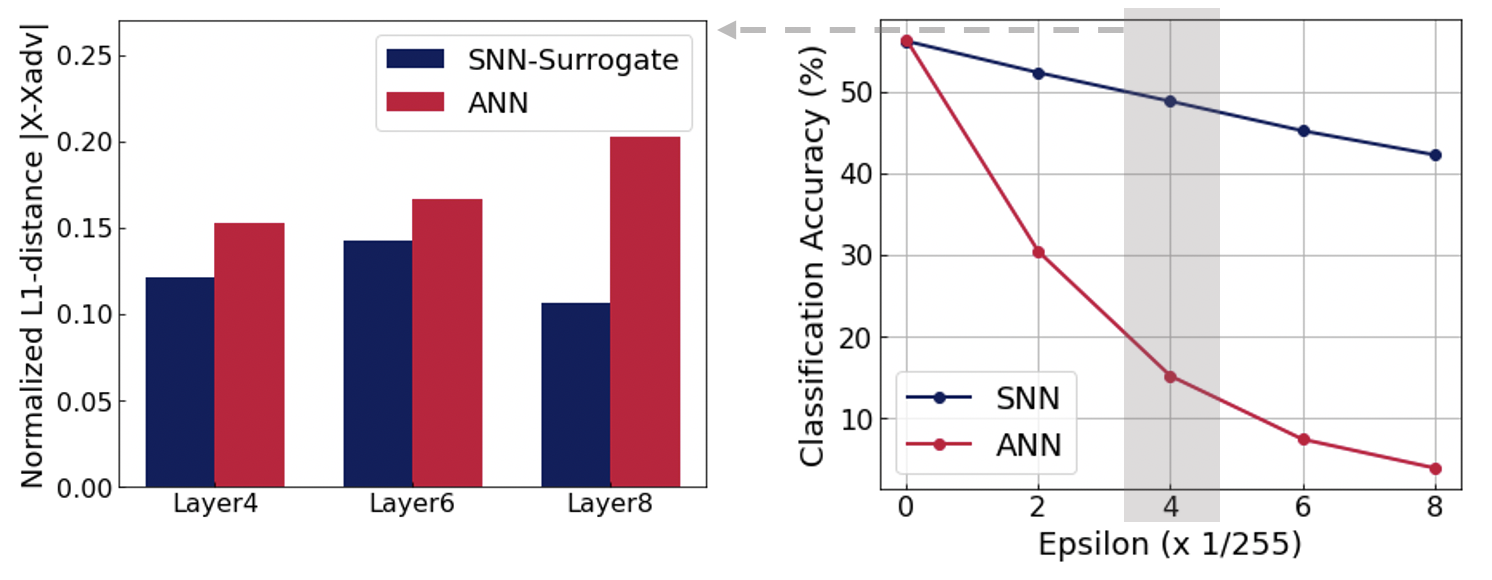}
    \vspace{1mm}
    \caption{Classification accuracy with respect to the FGSM intensity. We also compute the normalized L1 distance between heatmaps for clean and adversarial inputs at 
    $\epsilon = \frac{4}{255}$. 
    For SNN, we report the maximum difference among multiple time-steps.
    }
    \label{fig:fgsm_quantitative_result}
    \vspace{-3mm}
\end{figure}

\subsection{SAM: Unsupervised Visualization Tool}
\label{sec: Unsupervised Visualization Tool}
In Fig. \ref{fig:sam_grad_visualization}, we visualize the qualitative results of SAM on SNNs trained with surrogate learning as well as ANN-SNN conversion. We also show the Grad-CAM visualization obtained from a corresponding ANN for reference.
Note that SAM does not require any class label (unsupervised) compared to Grad-CAM that uses ground truth class labels.
Interestingly, heatmaps obtained from SAM across different time-steps on SNNs shows a similar  result with Grad-CAM on ANNs where the region of interest is highlighted in a discriminative fashion.
This  supports our assertion that SAM is an  effective  visualization  tool  for  SNNs.  
Moreover, the results imply that ISI and temporal dynamics can yield intepretability for deep SNNs.

\subsection{Surrogate Gradient Learning vs. Conversion}

We compare the SAM visualization results of surrogate gradient learning (Fig. \ref{fig:sam_grad_visualization}(c)) and conversion (Fig. \ref{fig:sam_grad_visualization}(d)).
From the figure, we observe a trend in the heatmap visualization of surrogate gradient learning with zero activity at early time-steps leading to discriminative activity in the mid-range followed by zero activity again towards the end.
In contrast, conversion maintains similar heatmaps during  the  entire  time  period. This is related to the variation in spike activity for each time-step as shown in Fig. \ref{fig:loc_comp}(b).
Since surrogate gradient learning considers a
temporal dynamic during training \cite{kim2020revisiting,roy2019towards},  each layer passes the information (\ie, the number of spikes) consecutively.
On the other hand, conversion does not show any temporal propagation. 
Moreover, we observe that surrogate gradient learning has more accurate (\ie similar to Grad-CAM from ANN) heatmaps highlighting the region of interest across all layers.
Notably, the conversion method highlights only partial regions of the object (\eg, lemon) and in some cases (\eg, bird) the wrong region.
This observation is supported by the localization error comparison in Fig. \ref{fig:loc_comp}(a). For all layers, surrogate gradient learning shows lower localization error. {It is well known and evident that conversion methods do not account for any temporal dynamics during training \cite{roy2019towards}. We believe that this missing temporal dependence accounts for less interpretability.} Thus, we assert that SNNs obtained with surrogate gradient learning (incorporating temporal dynamics) are more interpretable.  Therefore, all visualization analyses in the next sections focus on the surrogate gradient learning method.

\subsection{Intermediate Layers of SNN}
So far, no studies have analysed the underlying information learnt in different layers of an SNN. It has been always assumed that SNNs like ANNs learn features in a generic-to-specific manner as we go deeper. For the first time, we visualize the explanations at intermediate layers of SNN using SAM to support this assumption.
In Fig. \ref{fig:sam_grad_visualization} (see SAM-Surrogate results), the SAM visualization shows that shallow layers of SNNs represent low-level structure  and deep layers focus on semantic information.
For example, layer 4 highlights the edges or blobs of the lion, such as eyes and nose. On the other hand, layer 8 highlights the full face of the lion.

\begin{table}[t]
\addtolength{\tabcolsep}{1.5pt}
\centering
\caption{Ablation studies on leak factor $\lambda$. We show localization error and classification accuracy with respect to $\lambda$.}
\vspace{1mm}
\resizebox{0.4\textwidth}{!}
{
\begin{tabular}{lcccccccc}
\toprule
   & $\lambda = 0.7$  & $\lambda = 0.8$ & $\lambda = 0.9$ \\
\midrule
    Localization Error (Layer 4) & 0.68  & 0.63 & 0.61 \\
    Localization Error (Layer 6) & 0.86 & 0.82 & 0.80 \\
    Localization Error (Layer 8)  & 0.98 & 0.97 & 0.91  \\
\midrule
    Accuracy (\%) & 47.53 & 53.36 & 56.57  \\
\bottomrule
\end{tabular}%
}
\label{table: leak_ablation}
\vspace{-4mm}
\end{table}

\subsection{Effect of Leak in SNN}

We analyze the effect of leak factor $\lambda$ (Eq. \ref{eq:LIF}), one of the important parameters in SNNs.
The leak parameter $\lambda$ ($0<\lambda \leq 1$) controls the forgetting behavior of LIF neurons similar to the human brain.
We note that high $\lambda$ means less forgetting.
In order to explain the effect of leak on visualization, we measure the localization error for different leak values [0.7, 0.8, 0.9].
Table \ref{table: leak_ablation} shows that high leak parameter $\lambda$ achieves low localization error.
This is because a low $\lambda$ forgets the stored voltage in a neuron (\ie information) within a few time-steps and thus cannot produce any reasinable spike activity or visualization. 
We also compute the classification accuracy on Tiny-ImageNet in Table \ref{table: leak_ablation}. 
The results show that low $\lambda$ induces a drastic accuracy drop due to the excessive forgetting behavior. 
Overall, appropriate leak selection is important to achieve accurate localization/visualization as well as performance.

\subsection{Effect of Hyperparamter $\gamma$ }
\vspace{-1.5mm}

We conduct ablation studies to understand the effect of hyperparameter $\gamma$ on SAM in Eq. \ref{eq:tcsc}.
The $\gamma$ value decides the steepness of the exponential kernel function in TSCS.
A kernel with high $\gamma$ takes into account recent spike trajectory, where as low $\gamma$ considers long-period spike trajectory.
In Fig. \ref{fig:gamma_ablation}, we visualize the localization error with respect to $\gamma$ for different layers in VGG 11 for conversion and surrogate gradient methods.
For both methods, $\gamma = 0$ shows the highest localization error since the kernel does not filter redundant long ISI spikes.
Another interesting observation is that the localization error increases for large gamma value (\eg, 1.0). This is because high $\gamma$ limits reliable visualization by considering only very recent spikes and ignores spike history to a great extent. %Thus, we select $\gamma =0.2$ in our experiments to have maximum visualization. 

\vspace{-1mm}

\subsection{Adversarial Robustness of SNN}
\vspace{-1.5mm}

Previous studies \cite{sharmin2020inherent, sharmin2019comprehensive} have shown that SNNs are more robust to adversarial inputs than ANNs.
In order to observe the effectiveness of SNNs under attack, we conduct a qualitative and quantitative comparison between Grad-CAM and SAM.
We attack both ANN and SNN using FGSM attack \cite{goodfellow2014explaining} and SNN-crafted FGSM attack \cite{sharmin2020inherent} with  $\epsilon = \frac{4}{255}$.
In Fig. \ref{fig:sam_fgsm}, we can observe that Grad-CAM shows large change before/after attack. On the other hand, SAM shows almost similar results.
Moreover, we show the classification accuracy with respect to the attack intensity, and normalized L1-distance between heatmaps of clean and adversarial images at $\epsilon = \frac{4}{255}$, in Fig. \ref{fig:fgsm_quantitative_result}.
The results show that SNN is more robust than ANN in terms of both accuracy and visualization.
Therefore, using SNN with SAM for a secured system (\eg, military defense) will be a huge advantage in terms of robust interpretation.

\begin{figure}
    \centering
    \includegraphics[width=0.40\textwidth]{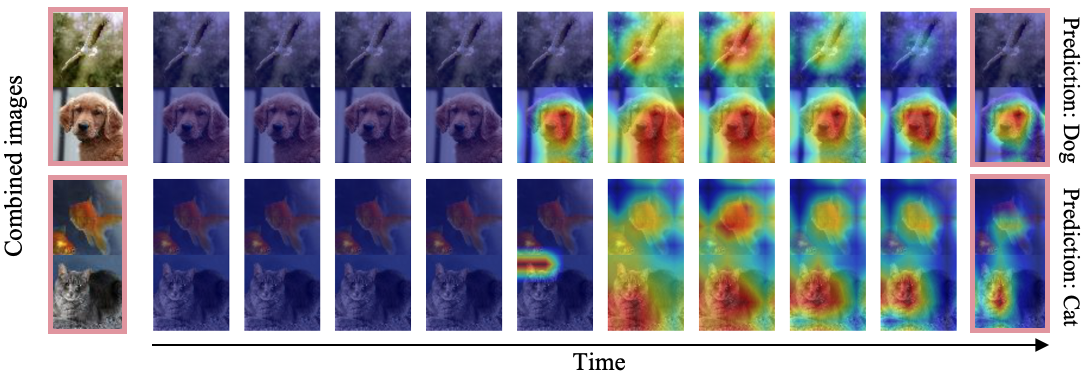}
    % \vspace{1mm}
    \caption{Visualization of SAM for multi-object images.
    }
    \label{fig:sensory_suppression}
    \vspace{-3.5mm}
\end{figure}

\subsection{Sensory Suppression Behavior of SNN}
\vspace{-1.5mm}

Neuroscience studies have suggested that human brain undergoes  \cite{kastner1998mechanisms,kastner2001neural,ungerleider2000mechanisms} ``sensory suppression". That is, the brain
focuses on one of multiple objects when these objects are presented at the same time.
%This is due to the limited processing capacity of the biological visual system \cite{ungerleider2000mechanisms}.
Co-incidentally, with SAM, we observe that SNNs also emulate sensory suppression when presented with multiple objects. 
%Since SNNs emulate the human brain, we implement the scenario of sensory suppression behavior in SNNs.
To show this, we concatenate two randomly chosen images from Tiny ImageNet dataset and pass the concatenated image into the SNN trained with surrogate gradient learning.
Interestingly, as shown in Fig. \ref{fig:sensory_suppression}, the results show that neurons compete in the earlier time-steps for attending to both objects and finally focus/attend on only one of the objects at later time-steps. Note, for each image, the final prediction from the SNN matches the final attention shown by SAM. These results unleash the bio-plausible characteristics of SNNs and also establish SAM as a suitable interpretation tool.

\vspace{-1.5mm}

\section{Conclusion}
\vspace{-1.5mm}
In this paper, we propose a visualization tool for SNNs, called SAM.
This is different from a conventional ANN visualization tool since SAM does not require any target labels and backpropagated gradients.
Instead, we use the temporal dynamics of SNNs to compute a neuronal contribution score in forward propagation based on the history of previous spikes.
Without any label, SAM highlight the discriminative region for prediction.
Through extensive experiments, we show the functionality of SAM in
various configuration of SNNs. Overall, SAM opens up the possibility towards interpretable neuromorphic computing.

\section{Acknowledgement}
The research was funded in part by C-BRIC, one of six centers in JUMP, a Semiconductor Research Corporation (SRC) program sponsored by DARPA, the National Science Foundation (Grant\#1947826), and the Amazon Research Award.

% \clearpage

{\small
\bibliographystyle{ieee_fullname}
\bibliography{egbib}
}

\end{document}